\documentclass{article}

\usepackage{times}
\usepackage{graphicx} 
\usepackage{caption}
\usepackage{subcaption}

\usepackage{amsmath}
\usepackage{amsfonts}
\usepackage{amssymb}
\usepackage{amsthm}

\usepackage[utf8]{inputenc}
\usepackage{amsthm}
\usepackage{amsmath}
\usepackage{amsfonts}
\usepackage{amssymb}
\usepackage{dsfont}
\usepackage{color}
\allowdisplaybreaks

\newcommand{\N}{\mathcal{N}}
\newcommand{\cL}{\mathcal{L}}

\newcommand{\td}{\text{d}}

\newcommand{\x}{\mathbf{x}}
\newcommand{\Bb}{\mathbf{b}}
\newcommand{\sBb}{\mathtt{z}}

\newcommand{\y}{\mathbf{y}}

\newcommand{\w}{\mathbf{w}}
\newcommand{\W}{\mathbf{W}}

\newcommand{\m}{\mathbf{m}}

\newcommand{\X}{\mathbf{X}}
\newcommand{\Y}{\mathbf{Y}}

\newcommand{\I}{\mathbf{I}}
\newcommand{\M}{\mathbf{M}}
\newcommand{\Mh}{\mathbf{W}}

\newcommand{\bo}{\text{\boldmath$\omega$}}

\newcommand{\K}{\mathbf{K}}

\newcommand{\Var}{\text{Var}}

\newcommand{\diag}{\text{diag}}
\newcommand{\KL}{\text{KL}}

\newcommand{\weightdecay}{\lambda}

\theoremstyle{definition}

\def\fl[#1\]{\begin{align}#1\end{align}}
\def\[#1\]{\begin{align*}#1\end{align*}}

\def\*[#1\]{\begin{align*}#1\end{align*}}

\usepackage{natbib}

\usepackage{algorithm}
\usepackage{algorithmic}

\usepackage{hyperref}

\usepackage[accepted]{icml2016}

\icmltitlerunning{Dropout as a Bayesian Approximation: 
Representing Model Uncertainty in Deep Learning}

\usepackage{dblfloatfix}

\begin{document} 

\twocolumn[
\icmltitle{Dropout as a Bayesian Approximation: \\
Representing Model Uncertainty in Deep Learning}

\icmlauthor{Yarin Gal}{yg279@cam.ac.uk}
\icmlauthor{Zoubin Ghahramani}{zg201@cam.ac.uk}
\icmladdress{University of Cambridge}

\icmlkeywords{machine learning, ICML, variational inference, Gaussian processes, dropout, deep learning, uncertainty}

]

\begin{abstract}
Deep learning tools have gained tremendous attention in applied machine learning. However such tools for regression and classification do not capture model uncertainty. In comparison, Bayesian models offer a mathematically grounded framework to reason about model uncertainty, but usually come with a prohibitive computational cost. In this paper we develop a new theoretical framework casting dropout training in deep neural networks (NNs) as approximate Bayesian inference in deep Gaussian processes. A direct result of this theory gives us tools to model uncertainty with dropout NNs -- extracting information from existing models that has been thrown away so far. This mitigates the problem of representing uncertainty in deep learning without sacrificing either computational complexity or test accuracy. We perform an extensive study of the properties of dropout's uncertainty. Various network architectures and non-linearities are assessed on tasks of regression and classification, using MNIST as an example. We show a considerable improvement in predictive log-likelihood and RMSE compared to existing state-of-the-art methods, and finish by using dropout's uncertainty in deep reinforcement learning. 
\vspace{-5mm}
\end{abstract} 

\section{Introduction}

Deep learning has attracted tremendous attention from researchers in fields such as physics, biology, and manufacturing, to name a few  \citep{baldi2014searching, anjos2015neural, bergmann2014use}. Tools such as neural networks (NNs), dropout, convolutional neural networks (convnets), and others are used extensively. However, these are fields in which representing model uncertainty is of crucial importance \citep{krzywinski2013points, ghahramani2015probabilistic}. With the recent shift in many of these fields towards the use of Bayesian uncertainty \citep{herzog2013experimental, Trafimow2015Editorial, nuzzo2014statistical}, new needs arise from deep learning tools. %

\begin{figure*}[t]
\vspace{-2mm}
	\centering
	\begin{subfigure}[b]{0.49\textwidth}
		\includegraphics[width=\linewidth, trim=2mm 3mm 2mm 2mm, clip]{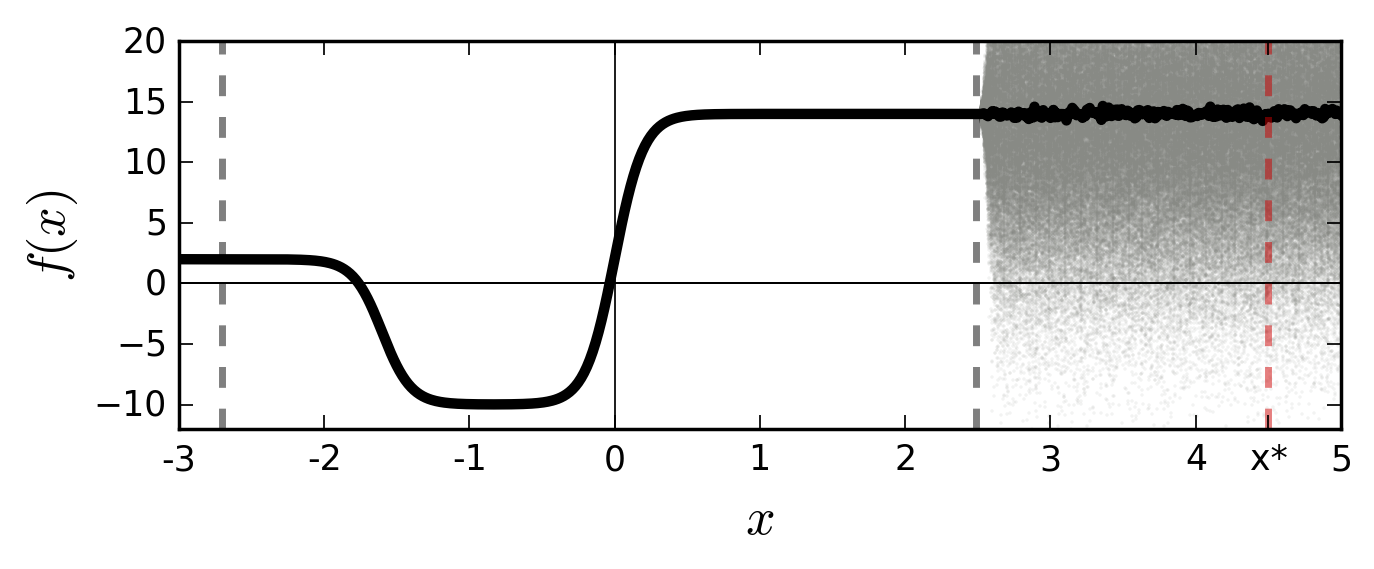}
		\vspace{-4mm}
		\caption{Arbitrary function $f(\x)$ as a function of data $\x$ (softmax \textit{input})} \label{fig:class_sketch_input}
	\end{subfigure}%
	\hspace{2mm}
	\begin{subfigure}[b]{0.49\textwidth}
		\vspace{1mm}
		\includegraphics[width=\linewidth, trim=2mm 3mm 2mm 2mm, clip]{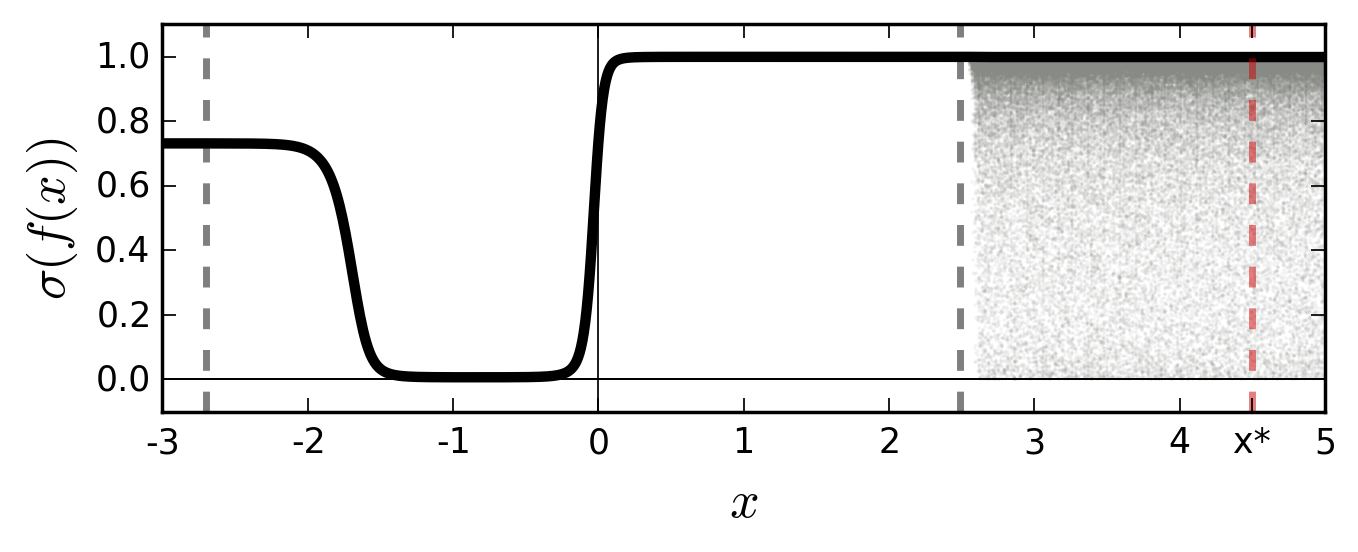}
		\vspace{-4mm}
		\caption{$\sigma(f(\x))$ as a function of data $\x$ (softmax \textit{output})} \label{fig:class_sketch_output}
	\end{subfigure}
	\vspace{-8mm}
	\caption{\textbf{A sketch of softmax input and output for an idealised binary classification problem.} Training data is given between the dashed grey lines. Function point estimate is shown with a solid line. Function uncertainty is shown with a shaded area. Marked with a dashed red line is a point $x^*$ far from the training data. Ignoring function uncertainty, point $x^*$ is classified as class 1 with probability $1$.} \label{fig:class_sketch}
\vspace{-4mm}
\end{figure*}

Standard deep learning tools for regression and classification do not capture model uncertainty. 
In classification, predictive probabilities obtained at the end of the pipeline (the softmax output) are often erroneously interpreted as model confidence. 
A model can be uncertain in its predictions even with a high softmax output (fig.\ \ref{fig:class_sketch}). 
Passing a point estimate of a function (solid line \ref{fig:class_sketch_input}) through a softmax (solid line \ref{fig:class_sketch_output}) results in extrapolations with unjustified high confidence for points far from the training data. $x^*$ for example would be classified as class 1 with probability $1$. However, passing the distribution (shaded area \ref{fig:class_sketch_input}) through a softmax (shaded area \ref{fig:class_sketch_output}) better reflects classification uncertainty far from the training data.
Model uncertainty is indispensable for the deep learning practitioner as well.
With model confidence at hand we can treat uncertain inputs and special cases explicitly. For example, in the case of classification, a model might return a result with high uncertainty. In this case we might decide to pass the input to a human for classification. This can happen in a post office, sorting letters according to their zip code, or in a nuclear power plant with a system responsible for critical infrastructure \citep{linda2009neural}.
Uncertainty is important in reinforcement learning (RL) as well \citep{szepesvari2010algorithms}. With uncertainty information an agent can decide when to exploit and when to explore its environment. Recent advances in RL have made use of NNs for Q-value function approximation. These are functions that estimate the quality of different actions an agent can take. Epsilon greedy search is often used where the agent selects its best action with some probability and explores otherwise. With uncertainty estimates over the agent's Q-value function, techniques such as Thompson sampling \citep{thompson1933likelihood} can be used to learn much faster. 

Bayesian probability theory offers us mathematically grounded tools to reason about model uncertainty, but these usually come with a prohibitive computational cost. 
It is perhaps surprising then that it is possible to 
cast recent deep learning tools as Bayesian models -- without changing either the models or the optimisation. 
We show that the use of dropout (and its variants) in NNs can be interpreted as a Bayesian approximation of a well known probabilistic model: the Gaussian process (GP) \citep{Rasmussen2005Gaussian}. Dropout is used in many models in deep learning as a way to avoid over-fitting \citep{srivastava2014dropout}, and our interpretation suggests that dropout approximately integrates over the models' weights. We develop tools for representing model uncertainty of existing dropout NNs -- extracting information that has been thrown away so far. This mitigates the problem of representing model uncertainty in deep learning without sacrificing either computational complexity or test accuracy.

In this paper we give a complete theoretical treatment of the link between Gaussian processes and dropout, and develop the tools necessary to represent uncertainty in deep learning. 
We perform an extensive exploratory assessment of the properties of the uncertainty obtained from dropout NNs and convnets on the tasks of regression and classification. We compare the uncertainty obtained from different model architectures and non-linearities in regression, and show that model uncertainty is indispensable for classification tasks, using MNIST as a concrete example. 
We then show a considerable improvement in predictive log-likelihood and RMSE compared to existing state-of-the-art methods.
Lastly we give a quantitative assessment of model uncertainty in the setting of reinforcement learning, on a practical task similar to that used in deep reinforcement learning \citep{mnih2015human}.\footnote{Code and demos are available at \url{http://yarin.co}.}

\section{Related Research}

It has long been known that infinitely wide (single hidden layer) NNs with distributions placed over their weights converge to Gaussian processes \citep{neal1995bayesian,williams1997computing}. 
This known relation is through a limit argument that does not allow us to translate properties from the Gaussian process to finite NNs easily. 
Finite NNs with distributions placed over the weights have been studied extensively as \textit{Bayesian neural networks} \citep{neal1995bayesian,mackay1992practical}. These offer robustness to over-fitting as well, but with challenging inference and additional computational costs. Variational inference has been applied to these models, but with limited success \citep{hinton1993keeping, barber1998ensemble, graves2011practical}. 
Recent advances in variational inference introduced new techniques into the field such as \textit{sampling-based} variational inference and \textit{stochastic} variational inference \citep{blei2012variational,kingma2013auto,
rezende2014stochastic,titsias2014doubly,
hoffman2013stochastic}. These have been used to obtain new approximations for Bayesian neural networks that perform as well as dropout \citep{blundell2015weight}. However these models come with a prohibitive computational cost. To represent uncertainty, the number of parameters in these models is doubled for the same network size. Further, they require more time to converge and do not improve on existing techniques. Given that good uncertainty estimates can be cheaply obtained from common dropout models, this might result in unnecessary additional computation. An alternative approach to variational inference makes use of expectation propagation \citep{hernandez2015probabilistic} and has improved considerably in RMSE and uncertainty estimation on VI approaches such as \citep{graves2011practical}. In the results section we compare dropout to these approaches and show a significant improvement in both RMSE and uncertainty estimation.

\section{Dropout as a Bayesian Approximation}

We show that a neural network with arbitrary depth and non-linearities, with dropout applied before every weight layer, is mathematically equivalent to an approximation to the probabilistic deep Gaussian process \citep{damianou2013deep} (marginalised over its covariance function parameters). We would like to stress that no simplifying assumptions are made on the use of dropout in the literature, and that the results derived are applicable to any network architecture that makes use of dropout exactly as it appears in practical applications. 
Furthermore, our results carry to other variants of dropout as well (such as drop-connect \citep{wan2013regularization}, multiplicative Gaussian noise \citep{srivastava2014dropout}, etc.).
We show that the dropout objective, in effect, minimises the Kullback--Leibler divergence between an approximate distribution and the posterior of a deep Gaussian process (marginalised over its finite rank covariance function parameters). 
Due to space constraints we refer the reader to the appendix for an in depth review of dropout, Gaussian processes, and variational inference (section 2), as well as the main derivation for dropout and its variations (section 3). The results are summarised here and in the next section we obtain uncertainty estimates for dropout NNs.

Let $\widehat{\y}$ be the output of a NN model with $L$ layers and a loss function $E(\cdot,\cdot)$ such as the softmax loss or the Euclidean loss (square loss). We denote by $\W_i$ the NN's weight matrices of dimensions $K_i \times K_{i-1}$, and by $\Bb_i$ the bias vectors of dimensions $K_i$ for each layer $i = 1, ..., L$. We denote by $\y_i$ the observed output corresponding to input $\x_i$ for $1 \leq i \leq N$ data points, and the input and output sets as $\X, \Y$.
During NN optimisation a regularisation term is often added.
We often use $L_2$ regularisation weighted by some weight decay $\weightdecay$, resulting in a minimisation objective (often referred to as cost),
\begin{align} \label{eq:L:dropout}
\cL_{\text{dropout}} := \frac{1}{N} \sum_{i=1}^N E(\y_i,\widehat{\y}_i) + \weightdecay \sum_{i=1}^L \big( ||\W_i||^2_2 + ||\Bb_i||^2_2 \big).
\end{align}
With dropout, we sample binary variables for every input point and for every network unit in each layer (apart from the last one). Each binary variable takes value 1 with probability $p_i$ for layer $i$. A unit is dropped (i.e.\ its value is set to zero) for a given input if its corresponding binary variable takes value 0. We use the same values in the backward pass propagating the derivatives to the parameters.

In comparison to the non-probabilistic NN, the deep Gaussian process 
is a powerful tool in statistics that allows us to model distributions over functions.
Assume we are given a covariance function of the form 
$$\K(\x, \y) = \int p(\w) p(b) \sigma(\w^T \x + b) \sigma(\w^T \y + b) \td \w \td b$$
 with some element-wise non-linearity $\sigma(\cdot)$ and distributions $p(\w), p(b)$. 
In sections 3 and 4 in the appendix we show that a deep Gaussian process with $L$ layers and covariance function $\K(\x, \y)$ can be approximated by placing a variational distribution over each component of a spectral decomposition of the GPs' covariance functions. This spectral decomposition maps each layer of the deep GP to a layer of explicitly represented hidden units, as will be briefly explained next.

Let $\Mh_i$ be a (now random) matrix of dimensions $K_i \times K_{i-1}$ for each layer $i$, and write $\bo = \{ \Mh_i \}_{i=1}^L$. A priori, we let each row of $\Mh_i$ distribute according to the $p(\w)$ above.
In addition, assume vectors $\m_i$ of dimensions $K_i$ for each GP layer. The predictive probability of the deep GP model (integrated w.r.t.\ the finite rank covariance function parameters $\bo$) given some precision parameter $\tau > 0$ can be parametrised as
\begin{align}
p(\y | \x, \X, \Y) &= \int p(\y | \x, \bo) p(\bo | \X, \Y) \td \bo \label{eq:predictive}\\
p(\y | \x, \bo) &= \N \big( \y; \widehat{\y}(\x, \bo), \tau^{-1} \I_D \big) \notag \\
\widehat{\y} \big(\x, \bo = \{\Mh_1, ..., &\Mh_L\} \big) \notag \\
&\hspace{-12mm} = \sqrt{\frac{1}{K_L}} \Mh_L \sigma \bigg( ... \sqrt{\frac{1}{K_1}} \Mh_2 \sigma \big( \Mh_1 \x + \m_1 \big) ... \bigg) \notag
\end{align}

The posterior distribution $p(\bo | \X, \Y)$ in eq.\ \eqref{eq:predictive} is intractable.
We use $q(\bo)$, a distribution over matrices whose columns are randomly set to zero, to approximate the intractable posterior. We define $q(\bo)$ as:
\begin{align*}
\Mh_i &= \M_i \cdot \diag([\sBb_{i,j}]_{j=1}^{K_i}) \\
\sBb_{i,j} &\sim \text{Bernoulli}(p_i) \text{ for } i = 1, ..., L, ~ j = 1, ..., K_{i-1}
\end{align*}
given some probabilities $p_i$ and matrices $\M_i$ as variational parameters. 
The binary variable $\sBb_{i,j} = 0$ corresponds then to unit $j$ in layer $i-1$ being dropped out as an input to layer $i$.
The variational distribution $q(\bo)$ is highly multi-modal, inducing strong joint correlations over the rows of the matrices $\Mh_i$ (which correspond to the frequencies in the sparse spectrum GP approximation).

We minimise the KL divergence between the approximate posterior $q(\bo)$ above and the posterior of the full deep GP, $p(\bo | \X, \Y)$.
This KL is our minimisation objective 
\begin{align}\label{eq:lowe_bound}
- \int q(\bo)  \log p(\Y|\X,\bo) \td \bo + \KL(q(\bo) || p(\bo)).
\end{align}
We rewrite the first term as a sum 
$$- \sum_{n=1}^N \int q(\bo) \log p(\y_n|\x_n,\bo) \td \bo$$ 
and approximate each term in the sum by Monte Carlo integration with a single sample $\widehat{\bo}_n \sim q(\bo)$ 
to get an unbiased estimate $- \log p(\y_n | \x_n, \widehat{\bo}_n)$. 
We further approximate the second term in eq.\ \eqref{eq:lowe_bound} and obtain $ \sum_{i=1}^L \big( \frac{p_i l^2}{2} ||\M_i||^2_2 + \frac{l^2}{2} ||\m_i||^2_2 \big)$ with prior length-scale $l$ (see section 4.2 in the appendix).
Given model precision $\tau$ we scale the result by the constant $1/\tau N$ to obtain the objective: 
\begin{align} \label{eq:L:GP-MC-reg}
\cL_{\text{GP-MC}} &\propto \frac{1}{N} \sum_{n=1}^N \frac{- \log p(\y_n | \x_n, \widehat{\bo}_n) }{\tau}\\
&\qquad + \sum_{i=1}^L \bigg( \frac{p_i l^2}{2 \tau N} ||\M_i||^2_2 + \frac{l^2}{2 \tau N} ||\m_i||^2_2 \bigg).\notag
\end{align}
Setting 
$$E(\y_n,\widehat{\y} (\x_n, \widehat{\bo}_n)) = - \log p(\y_n | \x_n, \widehat{\bo}_n) / \tau$$ 
we recover eq.\ \eqref{eq:L:dropout} for an appropriate setting of the precision hyper-parameter $\tau$ and length-scale $l$. The sampled $\widehat{\bo}_n$ result in realisations from the Bernoulli distribution $\sBb_{i,j}^n$ equivalent to the binary variables in the dropout case\footnote{In the appendix (section 4.1) we extend this derivation to classification. $E(\cdot)$ is defined as softmax loss and $\tau$ is set to 1.}.

\section{Obtaining Model Uncertainty}

We next derive results extending on the above showing that model uncertainty can be obtained from dropout NN models.

Following section 2.3 in the appendix, our approximate predictive distribution is given by
\begin{align} \label{eq:predictive_dist} 
q(\y^* | \x^*) = \int p(\y^* | \x^*, \bo) q(\bo) \td \bo
\end{align}
where $\bo = \{ \Mh_{i} \}_{i=1}^{L}$ is our set of random variables for a model with $L$ layers. 

We will perform moment-matching and estimate the first two moments of the predictive distribution empirically. More specifically, we sample $T$ sets of vectors of realisations from the Bernoulli distribution $\{ \sBb_{1}^t, ..., \sBb_{L}^t \}_{t=1}^T$ with $\sBb^t_{i} = [\sBb^t_{i,j}]_{j=1}^{K_i}$, 
giving $\{ \W_{1}^t, ..., \W_{L}^t \}_{t=1}^T$. We estimate
\begin{align} \label{eq:predictive_mean}
\mathbb{E}_{q(\y^* | \x^*)} (\y^*) \approx \frac{1}{T} \sum_{t=1}^T \widehat{\y}^*(\x^*, \Mh_{1}^t, ..., \Mh_{L}^t)
\end{align}
following proposition C in the appendix.
We refer to this Monte Carlo estimate as \textit{MC dropout}. In practice this is equivalent to performing $T$ stochastic forward passes through the network and averaging the results.

This result has been presented in the literature before as model averaging. We have given a new derivation for this result which allows us to derive mathematically grounded uncertainty estimates as well. 
\citet[][section 7.5]{srivastava2014dropout} have reasoned empirically that MC dropout can be approximated by averaging the weights of the network (multiplying each $\W_i$ by $p_i$ at test time, referred to as \textit{standard dropout}). 

We estimate the second raw moment in the same way:
\begin{align*}
&\mathbb{E}_{q(\y^* | \x^*)} \big( (\y^*)^T(\y^*) \big) \approx \tau^{-1} \I_D \\
&\qquad + \frac{1}{T} \sum_{t=1}^T \widehat{\y}^*(\x^*, \Mh_{1}^t, ..., \Mh_{L}^t)^T \widehat{\y}^*(\x^*, \Mh_{1}^t, ..., \Mh_{L}^t)
\end{align*}
following proposition D in the appendix.
To obtain the model's predictive variance we have:
\begin{align*}
&\Var_{q(\y^* | \x^*)} \big( \y^* \big) \approx \tau^{-1} \I_D \\
&\qquad + \frac{1}{T} \sum_{t=1}^T \widehat{\y}^*(\x^*, \Mh_{1}^t, ..., \Mh_{L}^t)^T \widehat{\y}^*(\x^*, \Mh_{1}^t, ..., \Mh_{L}^t)
\\
&\qquad - \mathbb{E}_{q(\y^* | \x^*)} (\y^*)^T \mathbb{E}_{q(\y^* | \x^*)} (\y^*)
\end{align*}
which equals the sample variance of $T$ stochastic forward passes through the NN plus the inverse model precision. Note that $\y^*$ is a row vector thus the sum is over the \textit{outer-products}.
Given the weight-decay $\weightdecay$ (and our prior length-scale $l$) we can find the model precision from the identity 
\begin{align} \label{eq:tau_weightdecay}
\tau = \frac{p l^2}{2 N \weightdecay}.
\end{align}

We can estimate our predictive log-likelihood by Monte Carlo integration of eq.\ \eqref{eq:predictive}. This is an estimate of how well the model fits the mean and uncertainty (see section 4.4 in the appendix). For regression this is given by:
\newcommand{\logsumexp}{\text{logsumexp}}
\begin{align} \label{eq:predictive_dist_reg}
\log p(\y^* | \x^*, \X, \Y) &\approx \logsumexp \bigg( -\frac{1}{2} \tau || \y - \widehat{\y}_t ||^2 \bigg) \notag \\
&\qquad - \log T - \frac{1}{2} \log 2 \pi - \frac{1}{2} \log \tau^{-1}
\end{align}
with a log-sum-exp of $T$ terms and $\widehat{\y}_t$ stochastic forward passes through the network. 

Our predictive distribution $q(\y^* | \x^*)$ is expected to be highly multi-modal, and the above approximations only give a glimpse into its properties. This is because the approximating variational distribution placed on each weight matrix column is bi-modal, and as a result the joint distribution over each layer's weights is multi-modal (section 3.2 in the appendix).

Note that the dropout NN model itself is not changed. To estimate the predictive mean and predictive uncertainty we simply collect the results of stochastic forward passes through the model. As a result, this information can be used with existing NN models trained with dropout. Furthermore, the forward passes can be done concurrently, resulting in constant running time identical to that of standard dropout. 

\begin{figure*}[t!]
\vspace{-2mm}
	\centering
	\begin{subfigure}[b]{0.5\textwidth}
		\includegraphics[width=\linewidth]{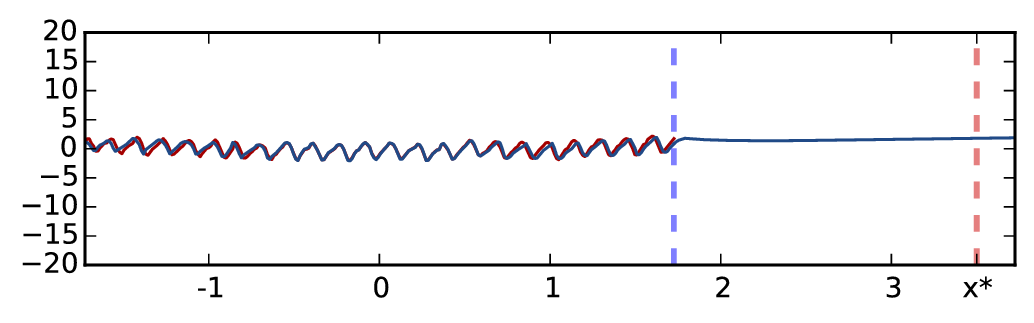}
		\vspace{-5mm}
		\caption{Standard dropout with weight averaging} \label{fig:extra_dropout}
	\end{subfigure}%
	\begin{subfigure}[b]{0.5\textwidth}
		\includegraphics[width=\linewidth]{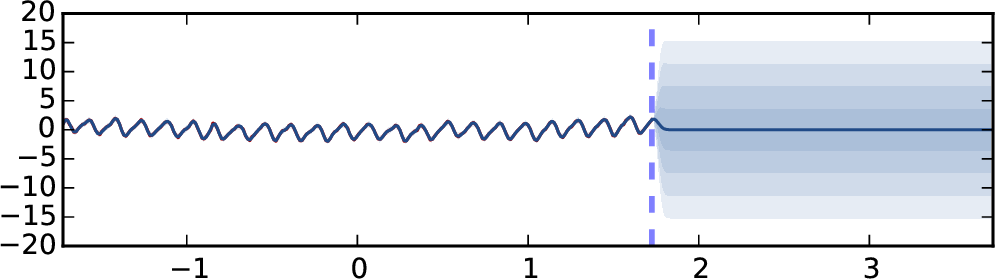}
		\vspace{-5mm}
		\caption{Gaussian process with SE covariance function} \label{fig:extra_GP}
	\end{subfigure}
	\begin{subfigure}[b]{0.5\textwidth}
		\includegraphics[width=\linewidth]{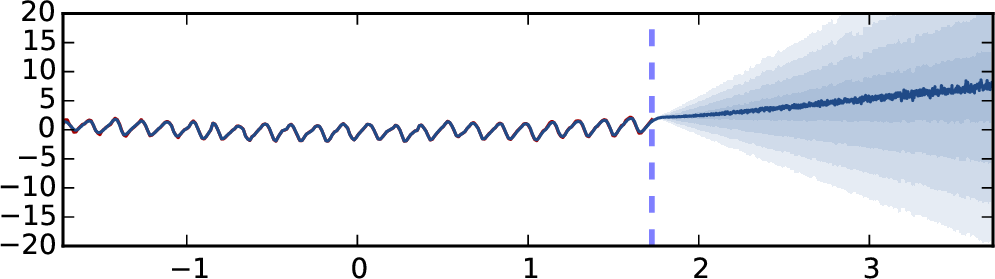}
		\vspace{-5mm}
		\caption{MC dropout with ReLU non-linearities} \label{fig:extra_MC_RELU}
	\end{subfigure}%
	\begin{subfigure}[b]{0.5\textwidth}
		\includegraphics[width=\linewidth]{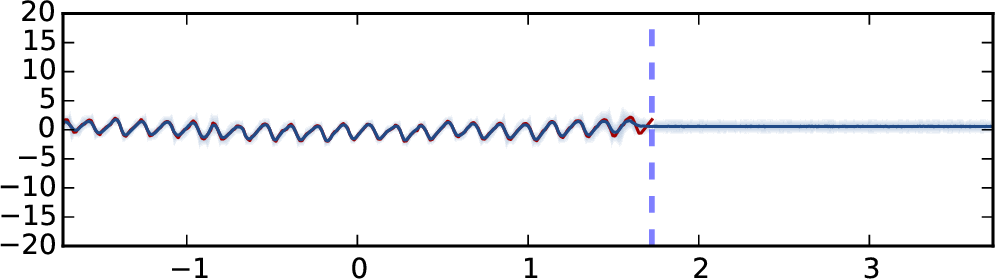}
		\vspace{-5mm}
		\caption{MC dropout with TanH non-linearities} \label{fig:extra_MC_TANH}
	\end{subfigure}
	\vspace{-6mm}
	\caption{\textbf{Predictive mean and uncertainties on the Mauna Loa CO$_2$ concentrations dataset, for various models.} In red is the observed function (left of the dashed blue line); in blue is the predictive mean plus/minus two standard deviations (8 for fig.\ \ref{fig:extra_MC_TANH}). Different shades of blue represent half a standard deviation. Marked with a dashed red line is a point far away from the data: standard dropout confidently predicts an insensible value for the point; the other models predict insensible values as well but with the additional information that the models are uncertain about their predictions.} \label{fig:extra}
	\vspace{-5mm}
\end{figure*}

\section{Experiments}

We next perform an extensive assessment of the properties of the uncertainty estimates obtained from dropout NNs and convnets on the tasks of regression and classification. We compare the uncertainty obtained from different model architectures and non-linearities, both on tasks of extrapolation, and show that model uncertainty is important for classification tasks using MNIST \citep{lecun1998mnist} as an example. 
We then show that using dropout's uncertainty we can obtain a considerable improvement in predictive log-likelihood and RMSE compared to existing state-of-the-art methods.
We finish with an example use of the model's uncertainty in a Bayesian pipeline. We give a quantitative assessment of the model's performance in the setting of reinforcement learning on a task similar to that used in deep reinforcement learning \citep{mnih2015human}.

Using the results from the previous section, we begin by qualitatively evaluating the dropout NN uncertainty on two regression tasks. We use two regression datasets and model scalar functions which are easy to visualise. These are tasks one would often come across in real-world data analysis. We use a subset of the atmospheric CO$_2$ concentrations dataset derived from in situ air samples collected at Mauna Loa Observatory, Hawaii \citep{Keeling2004} (referred to as \textit{CO$_2$}) to evaluate model extrapolation. In the appendix (section D.1) we give further results on a second dataset, the reconstructed solar irradiance dataset \citep{Lean2004}, to assess model interpolation. The datasets are fairly small, with each dataset consisting of about 200 data points. We centred and normalised both datasets.


\subsection{Model Uncertainty in Regression Tasks}

We trained several models on the CO$_2$ dataset. We use NNs with either 4 or 5 hidden layers and 1024 hidden units. We use either ReLU non-linearities or TanH non-linearities in each network, and use dropout probabilities of either $0.1$ or $0.2$. Exact experiment set-up is given in section E.1 in the appendix.

Extrapolation results are shown in figure \ref{fig:extra}. The model is trained on the training data (left of the dashed blue line), and tested on the entire dataset. Fig.\ \ref{fig:extra_dropout} shows the results for standard dropout (i.e.\ with weight averaging and without assessing model uncertainty) for the 5 layer ReLU model. Fig.\ \ref{fig:extra_GP} shows the results obtained from a Gaussian process with a squared exponential covariance function for comparison. Fig.\ \ref{fig:extra_MC_RELU} shows the results of the same network as in fig.\ \ref{fig:extra_dropout}, but with MC dropout used to evaluate the predictive mean and uncertainty for the training and test sets. Lastly, fig.\ \ref{fig:extra_MC_TANH} shows the same using the TanH network with 5 layers (plotted with 8 times the standard deviation for visualisation purposes). The shades of blue represent model uncertainty: each colour gradient represents half a standard deviation (in total, predictive mean plus/minus 2 standard deviations are shown, representing 95\% confidence). Not plotted are the models with 4 layers as these converge to the same results. 

Extrapolating the observed data, none of the models can capture the periodicity (although with a suitable covariance function the GP will capture it well). The standard dropout NN model (fig.\ \ref{fig:extra_dropout}) predicts value 0 for point $x^*$ (marked with a dashed red line) with high confidence, even though it is clearly not a sensible prediction. The GP model represents this by increasing its predictive uncertainty -- in effect declaring that the predictive value might be 0 but the model is uncertain. This behaviour is captured in MC dropout as well. Even though the models in figures \ref{fig:extra} have an incorrect predictive mean, the increased standard deviation expresses the models' uncertainty about the point. 

\begin{figure}[b!]
	\vspace{-6mm}
	\begin{minipage}{\linewidth}
		\includegraphics[width=\linewidth, height=1.75cm]{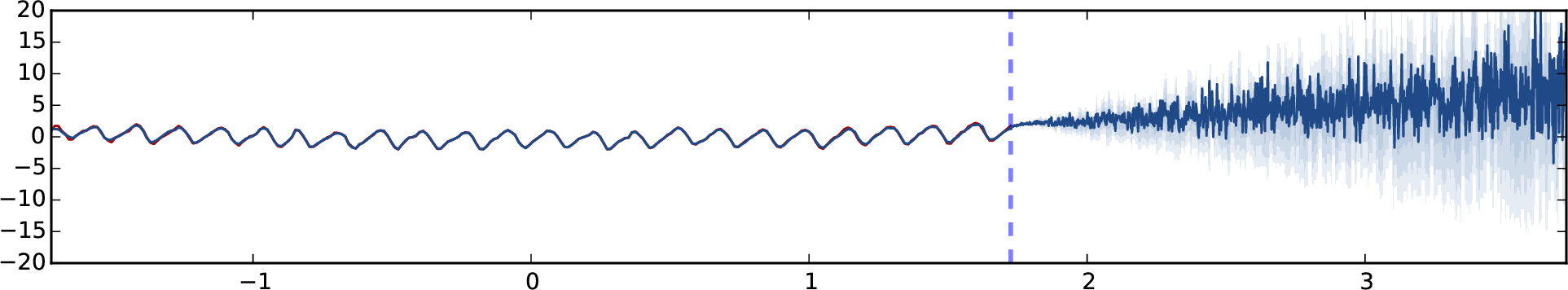}
		\vspace{-7mm}
		\caption{Predictive mean and uncertainties on the Mauna Loa CO$_2$ concentrations dataset for the MC dropout model with ReLU non-linearities, approximated with 10 samples.} \label{fig:MC_10}
	\end{minipage}
\end{figure}

Note that the uncertainty is increasing far from the data for the ReLU model, whereas for the TanH model it stays bounded. 

\begin{figure*}[t]
\vspace{-2mm}
	\centering
	\begin{subfigure}[b]{0.5\textwidth}
		\includegraphics[width=\linewidth, trim=0mm 0mm 0mm 0mm, clip]{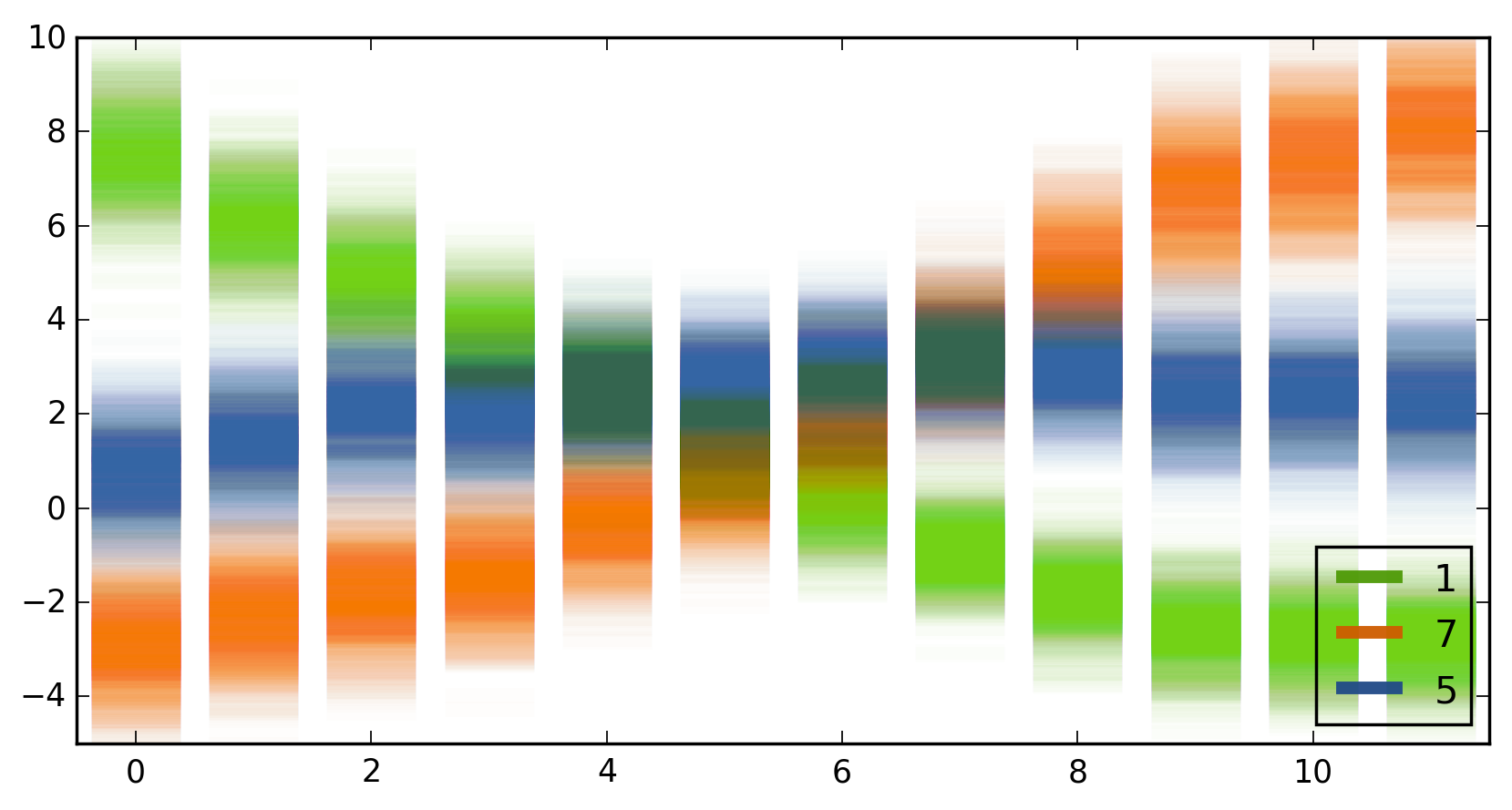}
		
		\vspace{-3.5mm}
		\hspace{2.5mm}
		\includegraphics[width=0.955\linewidth, trim=7mm 7mm 0mm 4mm, clip]{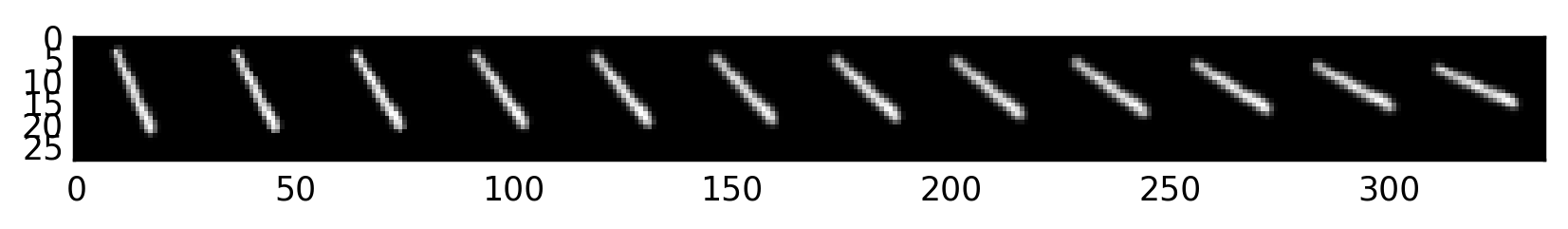}
		
		\caption{Softmax \textit{input} scatter} \label{fig:class_softmax_input}
	\end{subfigure}%
	\begin{subfigure}[b]{0.5\textwidth}
		\includegraphics[width=\linewidth, trim=0mm 0mm 0mm 0mm, clip]{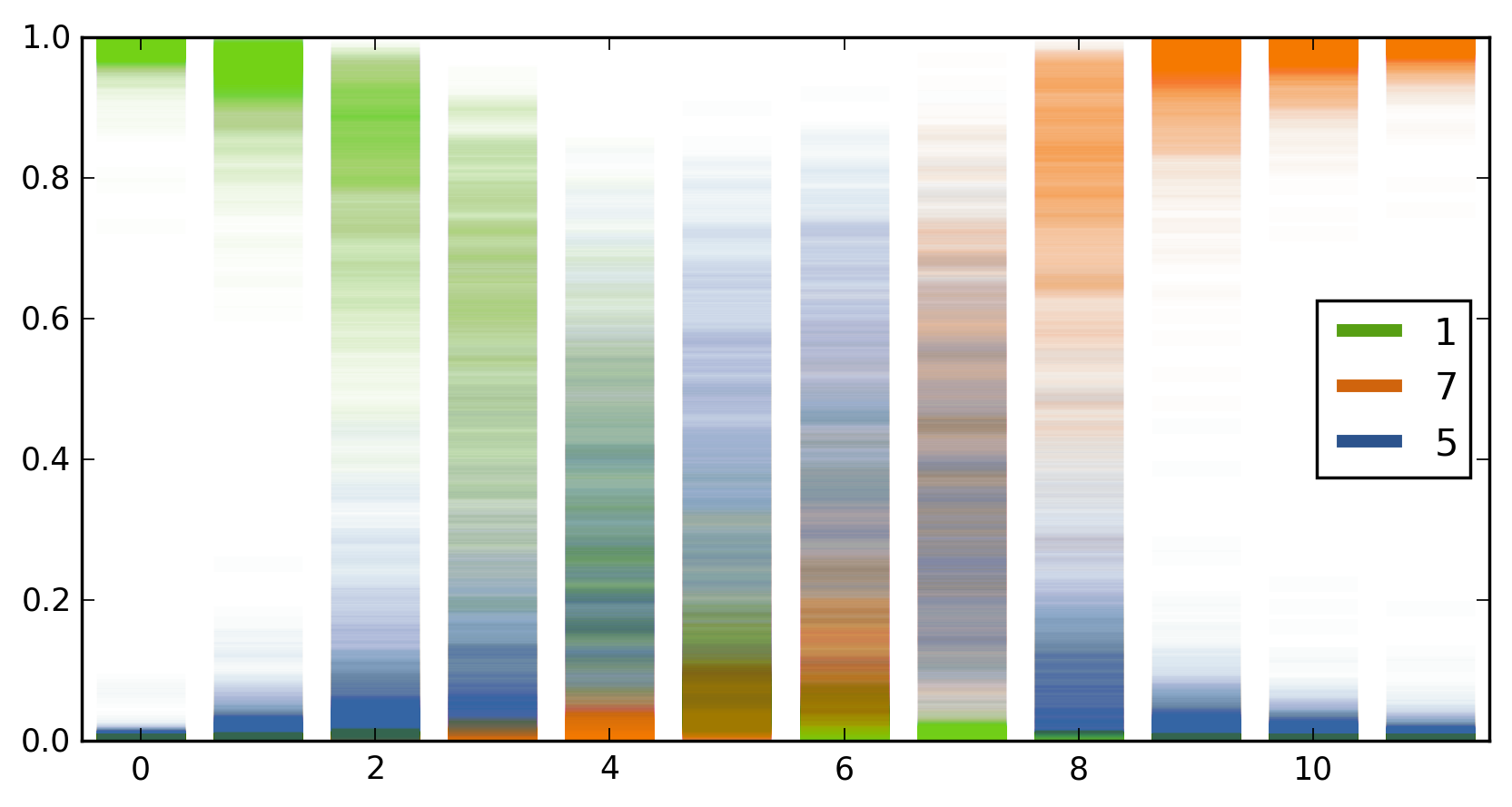}
		
		\vspace{-3.5mm}
		\hspace{2.5mm}
		\includegraphics[width=0.955\linewidth, trim=7mm 7mm 0mm 4mm, clip]{figs/exp2/MC_dropout_mnist_x_axis}
		
		\caption{Softmax \textit{output} scatter} \label{fig:class_softmax_output}
	\end{subfigure}
	\caption{\textbf{A scatter of 100 forward passes of the softmax input and output for dropout LeNet.} On the $X$ axis is a rotated image of the digit 1. The input is classified as digit 5 for images 6-7, even though model uncertainty is extremly large (best viewed in colour).} \label{fig:class_softmax}
	\vspace{-4mm}
\end{figure*}
This is not surprising, as dropout's uncertainty draws its properties from the GP in which different covariance functions correspond to different uncertainty estimates.
ReLU and TanH approximate different GP covariance functions (section 3.1 in the appendix) and TanH saturates whereas ReLU does not. 
For the TanH model we assessed the uncertainty using both dropout probability $0.1$ and dropout probability $0.2$. Models initialised with dropout probability $0.1$ initially exhibit smaller uncertainty than the ones initialised with dropout probability $0.2$, but towards the end of the optimisation when the model has converged the uncertainty is almost indistinguishable. It seems that the moments of the dropout models converge to the moments of the approximated GP model -- its mean and uncertainty.
It is worth mentioning that we attempted to fit the data with models with a smaller number of layers unsuccessfully.

The number of forward iterations used to estimate the uncertainty ($T$) was $1000$ for drawing purposes. A much smaller numbers can be used to get a reasonable estimation to the predictive mean and uncertainty (see fig.\ \ref{fig:MC_10} for example with $T=10$).

\subsection{Model Uncertainty in Classification Tasks}

To assess model classification confidence in a realistic example we test a convolutional neural network trained on the full MNIST dataset \citep{lecun1998mnist}.
We trained the LeNet convolutional neural network model \citep{lecun1998gradient} with dropout applied before the last fully connected inner-product layer (the usual way dropout is used in convnets). We used dropout probability of $0.5$. We trained the model for $10^6$ iterations with the same learning rate policy as before with $\gamma=0.0001$ and $p=0.75$. We used Caffe \citep{jia2014caffe} reference implementation for this experiment.

We evaluated the trained model on a continuously rotated image of the digit 1 (shown on the $X$ axis of fig.\ \ref{fig:class_softmax}).
We scatter 100 stochastic forward passes of the softmax input (the output from the last fully connected layer, fig.\ \ref{fig:class_softmax_input}), as well as of the softmax output for each of the top classes (fig.\ \ref{fig:class_softmax_output}). For the 12 images, the model predicts classes [1 1 1 1 1 5 5 7 7 7 7 7].

\newcommand{\tpm}{$\pm$}
\begin{table*}[t!]
\center
\small
\begin{tabular}{@{}l@{\hspace{4mm}}l@{\hspace{3mm}}l@{\hspace{4mm}}r@{\hspace{2mm}}r@{\hspace{2mm}}r@{\hspace{4mm}}r@{\hspace{2mm}}r@{\hspace{2mm}}r
@{}}
\multicolumn{3}{c}{} & 
\multicolumn{3}{c}{\footnotesize Avg. Test RMSE and Std. Errors} & 
\multicolumn{3}{c}{\footnotesize Avg. Test LL and Std. Errors} \\ 
\textbf{Dataset} & $N$ & $Q$ & 
\multicolumn{1}{c}{\textbf{VI}} & 
\multicolumn{1}{c}{\textbf{PBP}} & 
\multicolumn{1}{c}{\textbf{Dropout}} & 
\multicolumn{1}{c}{\textbf{VI}} & 
\multicolumn{1}{c}{\textbf{PBP}} & 
\multicolumn{1}{c}{\textbf{Dropout}} \\ 
\hline 
Boston Housing & 506 & 13 &  
4.32 \tpm 0.29 &  3.01 \tpm 0.18 & \textbf{2.97 \tpm 0.19} & 
-2.90 \tpm 0.07 & -2.57 \tpm 0.09 & \textbf{-2.46 \tpm 0.06} \\ 
Concrete Strength & 1,030 & 8 & 
7.19 \tpm 0.12 & 5.67 \tpm 0.09 & \textbf{5.23 \tpm 0.12} &
-3.39 \tpm 0.02 & -3.16 \tpm 0.02 & \textbf{-3.04 \tpm 0.02} \\ 
Energy Efficiency & 768 & 8 & 
2.65 \tpm 0.08 & 1.80 \tpm 0.05 & \textbf{1.66 \tpm 0.04} &
-2.39 \tpm 0.03 & -2.04 \tpm 0.02 & \textbf{-1.99 \tpm 0.02} \\ 
Kin8nm & 8,192 & 8 & 
\textbf{0.10 \tpm 0.00} & \textbf{0.10 \tpm 0.00} & \textbf{0.10 \tpm 0.00} & 
0.90 \tpm 0.01 & 0.90 \tpm 0.01 & \textbf{0.95 \tpm 0.01} \\ 
Naval Propulsion & 11,934 & 16 & 
\textbf{0.01 \tpm 0.00} & \textbf{0.01 \tpm 0.00} & \textbf{0.01 \tpm 0.00} &
3.73 \tpm 0.12 & 3.73 \tpm 0.01 & \textbf{3.80 \tpm 0.01} \\ 
Power Plant & 9,568 & 4 & 
4.33 \tpm 0.04 & 4.12 \tpm 0.03 & \textbf{4.02 \tpm 0.04} &
-2.89 \tpm 0.01 & -2.84 \tpm 0.01 & \textbf{-2.80 \tpm 0.01} \\ 
Protein Structure & 45,730 & 9 & 
4.84 \tpm 0.03 & 4.73 \tpm 0.01 & \textbf{4.36 \tpm 0.01} &
-2.99 \tpm 0.01 & -2.97 \tpm 0.00 & \textbf{-2.89 \tpm 0.00} \\ 
Wine Quality Red & 1,599 & 11 & 
0.65 \tpm 0.01 & 0.64 \tpm 0.01 & \textbf{0.62 \tpm 0.01} &
-0.98 \tpm 0.01 & -0.97 \tpm 0.01 & \textbf{-0.93 \tpm 0.01} \\ 
Yacht Hydrodynamics & 308 & 6 & 
6.89 \tpm 0.67 & \textbf{1.02 \tpm 0.05} & 1.11 \tpm 0.09 &
-3.43 \tpm 0.16 & -1.63 \tpm 0.02 & \textbf{-1.55 \tpm 0.03} \\ 
Year Prediction MSD & 515,345 & 90 & 
 9.034 \tpm NA & 8.879 \tpm NA & \textbf{8.849 \tpm NA} & 
-3.622 \tpm NA & -3.603 \tpm NA & \textbf{-3.588  \tpm NA} \\ 
\hline 
\end{tabular} 
\caption{\textbf{Average test performance in RMSE and predictive log likelihood} for a popular variational inference method (VI, \citet{graves2011practical}), Probabilistic back-propagation (PBP, \citet{hernandez2015probabilistic}), and dropout uncertainty (Dropout). Dataset size ($N$) and input dimensionality ($Q$) are also given.}\label{table:lml}
\end{table*}

The plots show the softmax input value and softmax output value for the 3 digits with the largest values for each corresponding input. When the softmax input for a class is larger than that of all other classes (class 1 for the first 5 images, class 5 for the next 2 images, and class 7 for the rest in fig \ref{fig:class_softmax_input}), the model predicts the corresponding class. Looking at the softmax input values, if the uncertainty envelope of a class is far from that of other classes' (for example the left most image) then the input is classified with high confidence. 
On the other hand, if the uncertainty envelope intersects that of other classes (such as in the case of the middle input image), then even though the softmax output can be arbitrarily high (as far as 1 if the mean is far from the means of the other classes), the softmax output uncertainty can be as large as the entire space. This signifies the model's uncertainty in its softmax output value -- i.e.\ in the prediction. 
In this scenario it would not be reasonable to use probit to return class 5 for the middle image when its uncertainty is so high. One would expect the model to ask an external annotator for a label for this input.
Model uncertainty in such cases can be quantified by looking at the entropy or variation ratios of the model prediction. 

\subsection{Predictive Performance}

Predictive log-likelihood captures how well a model fits the data, with larger values indicating better model fit. Uncertainty quality can be determined from this quantity as well (see section 4.4 in the appendix). We replicate the experiment set-up in \citet{hernandez2015probabilistic} and compare the RMSE and predictive log-likelihood of dropout (referred to as ``Dropout'' in the experiments) to that of Probabilistic Back-propagation (referred to as ``PBP'', \citep{hernandez2015probabilistic}) and to a popular variational inference technique in Bayesian NNs (referred to as ``VI'', \citep{graves2011practical}). The aim of this experiment is to compare the uncertainty quality obtained from a \textit{naive} application of dropout in NNs to that of specialised methods developed to capture uncertainty. 

Following our Bayesian interpretation of dropout (eq.\ \eqref{eq:L:GP-MC-reg}) we need to define a prior length-scale, and find an optimal model precision parameter $\tau$ which will allow us to evaluate the predictive log-likelihood (eq.\ \eqref{eq:predictive_dist_reg}).
Similarly to \citep{hernandez2015probabilistic} we use Bayesian optimisation (BO,  \citep{snoek2012practical,snoek2015spearmint}) over validation log-likelihood to find optimal $\tau$, and set the prior length-scale to $10^{-2}$ for most datasets based on the range of the data.
Note that this is a standard dropout NN, where the prior length-scale $l$ and model precision $\tau$ are simply used to define the model's weight decay through eq.\ \eqref{eq:tau_weightdecay}.
We used dropout with probabilities $0.05$ and $0.005$ since the network size is very small (with 50 units following \citep{hernandez2015probabilistic}) and the datasets are fairly small as well. 
The BO runs used 40 iterations following the original setup, but after finding the optimal parameter values we used 10x more iterations, as dropout takes longer to converge. Even though the model doesn't converge within 40 iterations, it gives BO a good indication of whether a parameter is good or not. Finally, we used mini-batches of size 32 and the Adam optimiser \citep{kingma2014adam}. 
Further details about the various datasets are given in \citep{hernandez2015probabilistic}.

The results are shown in table\footnote{\textbf{Update} [October 2016]: Note that in an earlier version of this paper our reported dropout standard error was erroneously scaled-up by a factor of 4.5 (i.e.\ for Boston RMSE we reported standard error 0.85 instead of 0.19 for example).} \ref{table:lml}. Dropout significantly outperforms all other models both in terms of RMSE as well as test log-likelihood on all datasets apart from Yacht, for which PBP obtains better RMSE.
All experiments were averaged on 20 random splits of the data (apart from Protein for which only 5 splits were used and Year for which one split was used). 
The median for most datasets gives much better performance than the mean. For example, on the Boston Housing dataset dropout achieves median RMSE of 2.68 with an IQR interval of [2.45, 3.35] and predictive log-likelihood median of  -2.34 with IQR [-2.54, -2.29]. In the Concrete Strength dataset dropout achieves median RMSE of 5.15.

To implement the model we used Keras \citep{Chollet2015}, an open source deep learning package based on Theano \citep{bergstra+al:2010-scipy}. 
In \citep{hernandez2015probabilistic} BO for VI seems to require a considerable amount of additional time compared to PBP.
However our model's running time (including BO) is comparable to PBP's Theano implementation\footnote{
\textbf{Update} [October 2016]: In the results above we attempted to match PBP's run time (hence used only 10x more epochs compared to PBP's 40 epochs). Experimenting with 100x more epochs compared to PBP (10x more epochs compared to the results in table \ref{table:lml}) gives a considerable improvement both in terms of test RMSE as well as test log-likelihood over the results in table \ref{table:lml}. We further assessed a model with two hidden layers instead of one (using the same number of units for the second layer). Both experiments are shown in table \ref{table:lml2} at the end of this document.
}. On Naval Propulsion for example our model takes 276 seconds on average per split (start-to-finish, divided by the number of splits). 
With the optimal parameters BO found, model training took 95 seconds. This is in comparison to PBP's 220 seconds. 
For Kin8nm our model requires 188 seconds on average including BO, 65 seconds without, compared to PBP's 156 seconds. 

\begin{figure*}[b!]
\vspace{-4mm}
	\begin{minipage}{0.5\linewidth}
		\vspace{0mm}
		\includegraphics[width=\linewidth]{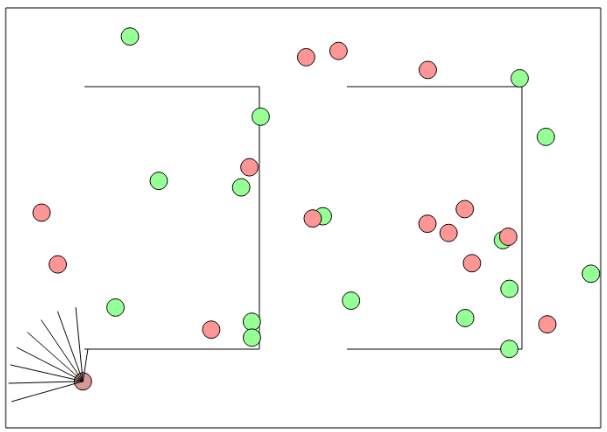}
		
		\vspace{1mm}
		\caption{Depiction of the reinforcement learning problem used in the experiments. The agent is in the lower left part of the maze, facing north-west.} \label{fig:rl_depict}
	\end{minipage}~~~
	\begin{minipage}{0.5\linewidth}
		\vspace{2mm}
		\includegraphics[width=\linewidth]{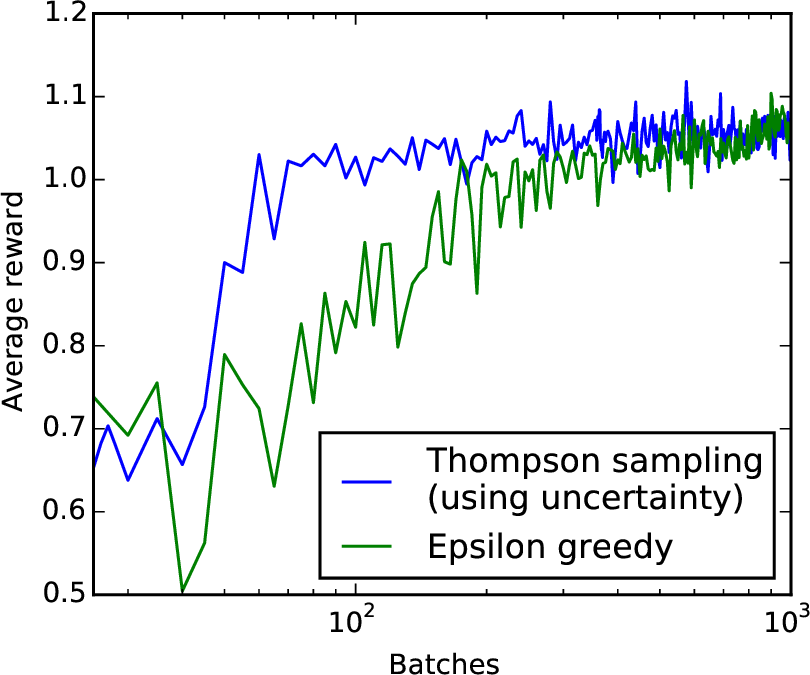}
		\vspace{-9mm}
		\caption{\textbf{Log plot of average reward} obtained by both epsilon greedy (in green) and our approach (in blue), as a function of the number of batches.} \label{fig:rl_plot}
	\end{minipage}
\end{figure*}

Dropout's RMSE in table \ref{table:lml} is given by averaging stochastic forward passes through the network following eq.\ \eqref{eq:predictive_mean} (MC dropout).
We observed an improvement using this estimate compared to the standard dropout weight averaging, and also compared to much smaller dropout probabilities (near zero). 
For the Boston Housing dataset for example, repeating the same experiment with dropout probability 0 results in RMSE of 3.07 and predictive log-likelihood of -2.59.
This demonstrates that dropout significantly affects the predictive log-likelihood and RMSE, even though the dropout probability is fairly small.

We used dropout following the same way the method would be used in current research -- without adapting model structure. This is to demonstrate the results that could be obtained from existing models when evaluated with MC dropout. Experimenting with different network architectures we expect the method to give even better uncertainty estimates.

\subsection{Model Uncertainty in Reinforcement Learning}

In reinforcement learning an agent receives various rewards from different states, and its aim is to maximise its expected reward over time.
The agent tries to learn to avoid transitioning into states with low rewards, and to pick actions that lead to better states instead.
Uncertainty is of great importance in this task -- with uncertainty information an agent can decide when to exploit rewards it knows of, and when to explore its environment. 

Recent advances in RL have made use of NNs to estimate agents' Q-value functions (referred to as Q-networks), a function that estimates the quality of different actions an agent can take at different states.
This has led to impressive results on Atari game simulations, where agents superseded human performance on a variety of games \citep{mnih2015human}. 
Epsilon greedy search was used in this setting, where the agent selects the best action following its current Q-function estimation with some probability, and explores otherwise. With our uncertainty estimates given by a dropout Q-network we can use techniques such as Thompson sampling \citep{thompson1933likelihood} to converge faster than epsilon greedy while avoiding over-fitting. 

We use code by \citep{convnetjs2015} that replicated the results by \citep{mnih2015human} with a simpler 2D setting. 
We simulate an agent in a 2D world with 9 eyes pointing in different angles ahead (depicted in fig.\ \ref{fig:rl_depict}). Each eye can sense a single pixel intensity of 3 colours. The agent navigates by using one of 5 actions controlling two motors at its base. An action turns the motors at different angles and different speeds. The environment consists of red circles which give the agent a positive reward for reaching, and green circles which result in a negative reward. The agent is further rewarded for not looking at (white) walls, and for walking in a straight line. 

We trained the original model, and an additional model with dropout with probability 0.1 applied before the every weight layer. Note that both agents use the same network structure in this experiment for comparison purposes. In a real world scenario using dropout we would use a larger model (as the original model was intentially selected to be small to avoid over-fitting).
To make use of the dropout Q-network's uncertainty estimates, we use Thompson sampling instead of epsilon greedy. In effect this means that we perform a single stochastic forward pass through the network every time we need to take an action. In replay, we perform a single stochastic forward pass and then back-propagate with the sampled Bernoulli random variables. Exact experiment set-up is given in section E.2 in the appendix.

In fig.\ \ref{fig:rl_plot} we show a \textit{log plot} of the average reward obtained by both the original implementation (in green) and our approach (in blue), as a function of the number of batches.
Not plotted is the burn-in intervals of 25 batches (random moves). 
Thompson sampling gets reward larger than $1$ within 25 batches from burn-in. Epsilon greedy takes 175 batches to achieve the same performance.
It is interesting to note that our approach seems to stop improving after 1K batches. This is because we are still sampling random moves, whereas epsilon greedy only exploits at this stage. 

\section{Conclusions and Future Research}

We have built a probabilistic interpretation of dropout which allowed us to obtain model uncertainty out of existing deep learning models. We have studied the properties of this uncertainty in detail, and demonstrated possible applications, interleaving Bayesian models and deep learning models together. 
This extends on initial research studying dropout from the Bayesian perspective \citep{Wang2013Fast,Maeda2014Bayesian}.

Bernoulli dropout is only one example of a regularisation technique corresponding to an approximate variational distribution which results in uncertainty estimates. Other variants of dropout 
follow our interpretation as well and correspond to alternative approximating distributions. These would result in different uncertainty estimates, trading-off uncertainty quality with computational complexity. We explore these in follow-up work.

Furthermore, each GP covariance function has a one-to-one correspondence with the combination of both NN non-linearities and weight regularisation. This suggests techniques to select appropriate NN structure and regularisation based on our a priori assumptions about the data. For example, if one expects the function to be smooth and the uncertainty to increase far from the data, cosine non-linearities and $L_2$ regularisation might be appropriate. The study of non-linearity--regularisation combinations and the corresponding predictive mean and variance are subject of current research.

\subsubsection*{Acknowledgements}
The authors would like to thank Dr Yutian Chen, Mr Christof Angermueller, Mr Roger Frigola, Mr Rowan McAllister, Dr Gabriel Synnaeve, Mr Mark van der Wilk, Mr Yan Wu, and many other reviewers for their helpful comments. Yarin Gal is supported by the Google European Fellowship in Machine Learning. 

\bibliography{NIPS_2015_deep_learning_uncertainty}
\bibliographystyle{icml2016}

\newpage
\appendix

\section{Appendix}
The appendix for the paper is given at \url{http://arxiv.org/abs/1506.02157}.

\begin{table*}[t!]
\center
\small
\begin{tabular}{@{}l@{\hspace{8mm}}r@{\hspace{2mm}}r@{\hspace{4mm}}r@{\hspace{8mm}}r@{\hspace{2mm}}r@{\hspace{4mm}}r
@{}}
\multicolumn{1}{c}{} & 
\multicolumn{3}{c}{\footnotesize Avg. Test RMSE and Std. Errors} & 
\multicolumn{3}{c}{\footnotesize Avg. Test LL and Std. Errors} \\ 
\textbf{Dataset} & 
\multicolumn{1}{l}{\textbf{Dropout}} & 
\multicolumn{1}{c}{\textbf{10x Epochs}} & 
\multicolumn{1}{l}{\textbf{2 Layers}} & 
\multicolumn{1}{l}{\textbf{Dropout}} & 
\multicolumn{1}{c}{\textbf{10x Epochs}} & 
\multicolumn{1}{c}{\textbf{2 Layers}} \\ 
\hline 
Boston Housing & 
2.97 \tpm 0.19 &  2.80 \tpm 0.19 & 2.80 \tpm 0.13 & 
-2.46 \tpm 0.06 & -2.39 \tpm 0.05 & -2.34 \tpm 0.02 \\ 
Concrete Strength & 
5.23 \tpm 0.12 & 4.81 \tpm 0.14 & 4.50 \tpm 0.18 &
-3.04 \tpm 0.02 & -2.94 \tpm 0.02 & -2.82 \tpm 0.02 \\ 
Energy Efficiency & 
1.66 \tpm 0.04 & 1.09 \tpm 0.05 & 0.47 \tpm 0.01 &
-1.99 \tpm 0.02 & -1.72 \tpm 0.02 & -1.48 \tpm 0.00 \\ 
Kin8nm & 
0.10 \tpm 0.00 & 0.09 \tpm 0.00 & 0.08 \tpm 0.00 & 
0.95 \tpm 0.01 & 0.97 \tpm 0.01 & 1.10 \tpm 0.00 \\ 
Naval Propulsion & 
0.01 \tpm 0.00 & 0.00 \tpm 0.00 & 0.00 \tpm 0.00 &
3.80 \tpm 0.01 & 3.92 \tpm 0.01 & 4.32 \tpm 0.00 \\ 
Power Plant & 
4.02 \tpm 0.04 & 4.00 \tpm 0.04 & 3.63 \tpm 0.04 &
-2.80 \tpm 0.01 & -2.79 \tpm 0.01 & -2.67 \tpm 0.01 \\ 
Protein Structure & 
4.36 \tpm 0.01 & 4.27 \tpm 0.01 & 3.62 \tpm 0.01 &
-2.89 \tpm 0.00 & -2.87 \tpm 0.00 & -2.70 \tpm 0.00 \\ 
Wine Quality Red & 
0.62 \tpm 0.01 & 0.61 \tpm 0.01 & 0.60 \tpm 0.01 &
-0.93 \tpm 0.01 & -0.92 \tpm 0.01 & -0.90 \tpm 0.01 \\ 
Yacht Hydrodynamics & 
1.11 \tpm 0.09 & 0.72 \tpm 0.06 & 0.66 \tpm 0.06 &
-1.55 \tpm 0.03 & -1.38 \tpm 0.01 & -1.37 \tpm 0.02 \\ 
\hline 
\end{tabular} 
\caption{\textbf{Average test performance in RMSE and predictive log likelihood} for dropout uncertainty as above (\textbf{Dropout}), the same model optimised with 10 times the number of epochs and identical model precision (\textbf{10x epochs}), and the same model again with 2 layers instead of 1 (\textbf{2 Layers}).}
\label{table:lml2}
\end{table*}

\end{document}